\documentclass[a4paper,10pt]{article}
\pdfoutput=1
\usepackage{nips12submit_e,times}
\usepackage{amsmath,amsfonts,amssymb,bbm,xspace,ulem,euscript,algorithm,algorithmic,graphicx,xcolor,subfigure,url,wrapfig,mathrsfs}

\newcommand{\x}{{\boldsymbol{x}}}
\renewcommand{\c}{{\boldsymbol{c}}}
\renewcommand{\r}{{\boldsymbol{r}}}
\newcommand{\transp}{^{\top}}
\newcommand{\eps}{\varepsilon}

\title{GRED: Graph-Regularized 3D Shape Reconstruction from Highly Anisotropic and Noisy Images}
\author{Christian Widmer\\
  Sloan-Kettering Institute \\
  1275 York avenue, New York, USA\\
  \texttt{cwidmer@cbio.mskcc.org}\\
  \And
  Philipp Drewe\\
  Sloan-Kettering Institute \\
  1275 York avenue, New York, USA\\
  \texttt{drewe@cbio.mskcc.org}\\
  \And
  Xinghua Lou\\
  Sloan-Kettering Institute \\
  1275 York avenue, New York, USA\\
  \texttt{loux@cbio.mskcc.org}\\
  \And
  Shefali Umrania\\
  Sloan-Kettering Institute \\
  1275 York avenue, New York, USA\\
  \texttt{umrania.shefali@gmail.com}\\
  \And
  Stephanie Heinrich\\
  Friedrich Miescher Laboratory\\
  Spemannstr. 39, T\"ubingen, Germany\\
  \texttt{stephanieheinrich@web.de}\\
  \And
  Gunnar R\"atsch\\
  Sloan-Kettering Institute \\
  1275 York avenue, New York, USA\\
  \texttt{raetsch@cbio.mskcc.org}}

\begin{document}
\nipsfinaltrue

\bibliographystyle{alpha}
\maketitle

\begin{abstract}
  Analysis of microscopy images can provide insight into many biological processes. 
  One particularly challenging problem is cell nuclear segmentation in highly anisotropic and noisy 3D image data. Manually localizing and segmenting each and every cell nuclei is very time consuming, which remains a bottleneck in large scale biological experiments.  In this work we
  present a tool for automated segmentation of  cell nuclei from 3D fluorescent microscopic data. Our tool is based on state-of-the-art image processing and machine learning techniques and supports a friendly graphical user interface (GUI). We show that our tool is as accurate as manual annotation but greatly
  reduces the time for the registration.\medskip

  {\bf Availability:} The software and data is available from \url{http://raetschlab.org/suppl/stef}.
% Cover the following:
% \begin{itemize}
% \item Motivation
% \item Why it is useful in biology
% \item Shortcomings of existing software solution
% \item Software availability
% \end{itemize}
%
\end{abstract}

\section{Introduction}
Imaging data, such as those form microscopic experiments, is a unique source of information in biology. Through fluorescent staining, they allow
to investigate tissue composition, cell shapes and  also sub-cellular
localization. A challenge however is that manual and consistent
measurements of such data is still time consuming and this remains an obstacle in  large scale experiments. Methods
that assist processing such complex, large data are therefore needed. These methods should not only speed up these
measurements steps but also increase the reproducibility of the measurements.  In this work we
will focus on the challenge of detecting cell nuclei from fluorescent microscopy images.  In
fluorescence microscopy, it is common practice for biologists to manually segment cells
based on 3D visualization and then later quantify the signal within this segmentation (usually
in a different staining channel). In particular, we will address nuclear segmentation for  anisotropic and high noisy 3D microscopic images with possible staining defect -- a very challenging problem that cannot be handled robustly by conventional computer vision methods such as blob detection, deformable model (e.g. level set) or combinational optimization (e.g. graph cut), because staining defect normally leads to missing intensity within the body of the true nuclear. A robust approach is highly desired. 

Though microscopes are sometimes equipped with software to assist researchers on this task;
more often than not the existing software only provides very rough polygon fits based on
intensity values. A major drawback of these existing software solutions is that 3D information is
not taken into account, which means segmentations are performed for each layer individually.
Furthermore, each object in a bigger volume (with potentially hundreds of cells) has to be processed
separately, making this step a major bottleneck.
Finally, any prior knowledge about the structure of the objects of interest is ignored as fits are
usually non-parametric. This is suboptimal when segmenting cells or nuclei, as these objects have
known structure that can be exploited.  Due to these drawbacks, signal quantification is an
extremely time consuming task, and truly large scale quantification experiments become prohibitive.
We propose a new method that addresses these shortcomings. It can be applied to images containing
multiple cells and exploits the fact that nuclei commonly have an ellipsoid shape.  The method
adapts graph-regularized transfer learning \cite{evgeniou2006learning} to the problem of parametric
fitting in several layers in combination with a robust loss function, as used in support vector
regression, to minimize the need for manual post-processing.  Our proposed method thus provides
biologists with a tool for high-throughput quantification experiments.

\section{Methods}

Our method performs the fitting in two steps: a preprocessing step to localize the nuclei that is
based on multi-scale Hessian eigenvalue thresholding~\cite{Lou2012Learning}, followed parametric
fitting procedure to compute the shape of each nucleus. These two steps will be explained in detail in
the following.

%The preprocessing step will 
%result in a set of sub-volumes that are subsequently processed by our fitting procedure. 
%The fitting procedure fits a stack of ellipses to each sub-volume, 
%such that robustness and smoothness and maximized. Our method combines
%ideas from support vector regression and transfer learning \cite{evgeniou2006learning}.
%In the following, we will carefully explain our method. We start by stating the simplest possible
%solution, which is fitting a circle to a number of points, which is then extended in a number
%of steps until we reach our final solution.

\subsection{Preprocessing}

For preprocessing, we localize and extract individual nuclei from a larger volume. We apply Hessian
eigenvalue thresholding introduced in \cite{Lou2012Learning}, which finds sets of foreground pixels
that cluster together.  For this we use Gaussian smoothing to aggregate mass from the neighborhood
and emphasize the central regions of the blobs (local maxima). A Hessian representation is then used
to find those local maxima, exploiting that
they have negative Hessian eigenvalues while a stationary point (e.g., a saddle point) does not. \cite{Gonzalez2008Digital}.
%\Phil{Still a bit hard to understand what is going on here}
This is one for multiple resolutions in
order to find clusters of all sizes.
\begin{figure}[H]
  \centering
  \subfigure[Raw image]{
    \label{preproc_0}
    \includegraphics[width=0.30\textwidth]{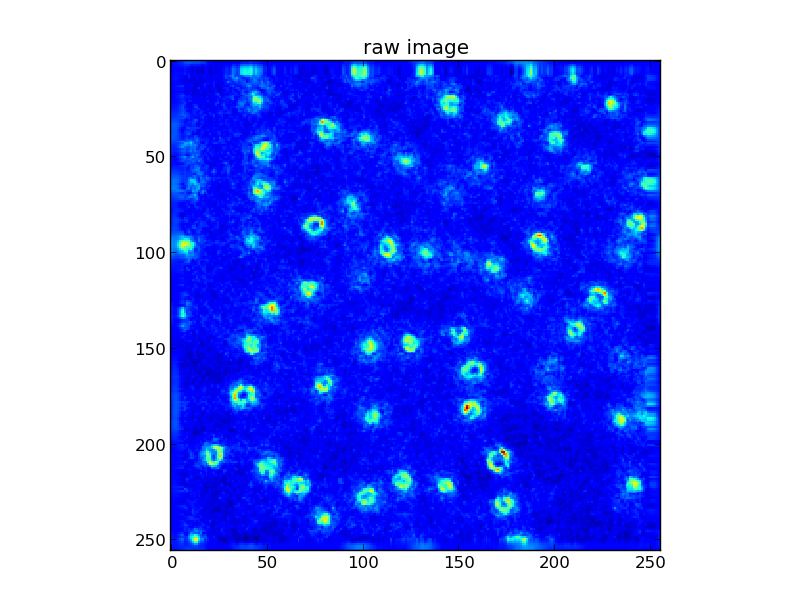}
  }
  \subfigure[Gaussian Smoothing]{
    \label{preproc_1a}
    \includegraphics[width=0.30\textwidth]{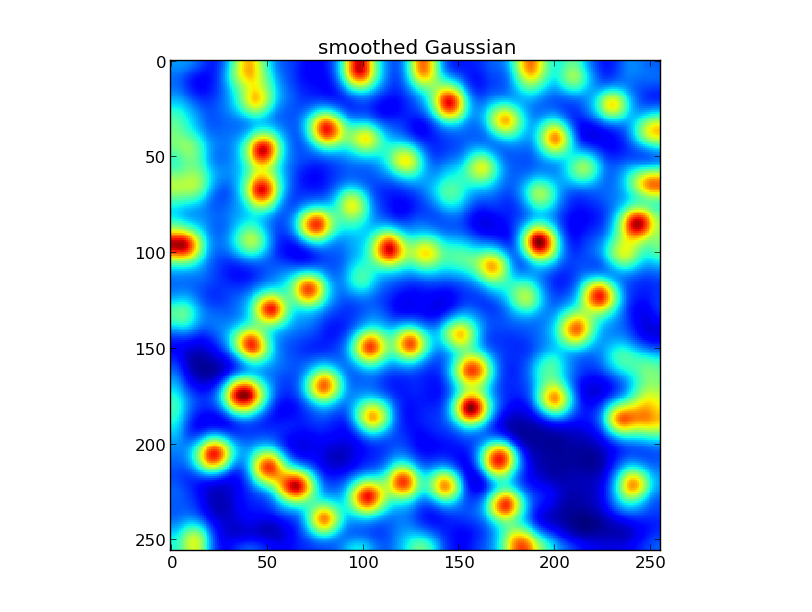}
  }
  \subfigure[Eigenvalues]{
    \label{preproc_1b}
    \includegraphics[width=0.30\textwidth]{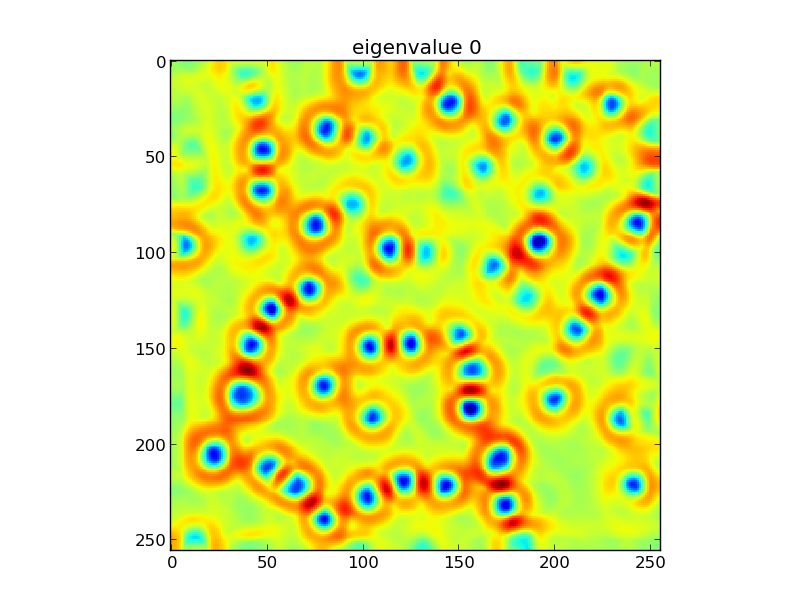}
  }
  \subfigure[Seeds]{
    \label{preproc_0a}
    \includegraphics[width=0.30\textwidth]{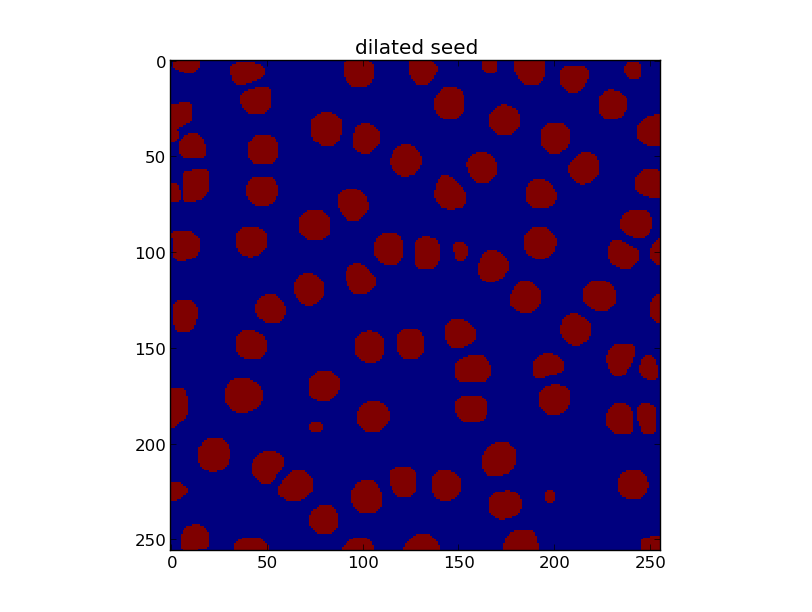}
  }
  \subfigure[Labeled Seeds]{
    \label{preproc_1c}
    \includegraphics[width=0.30\textwidth]{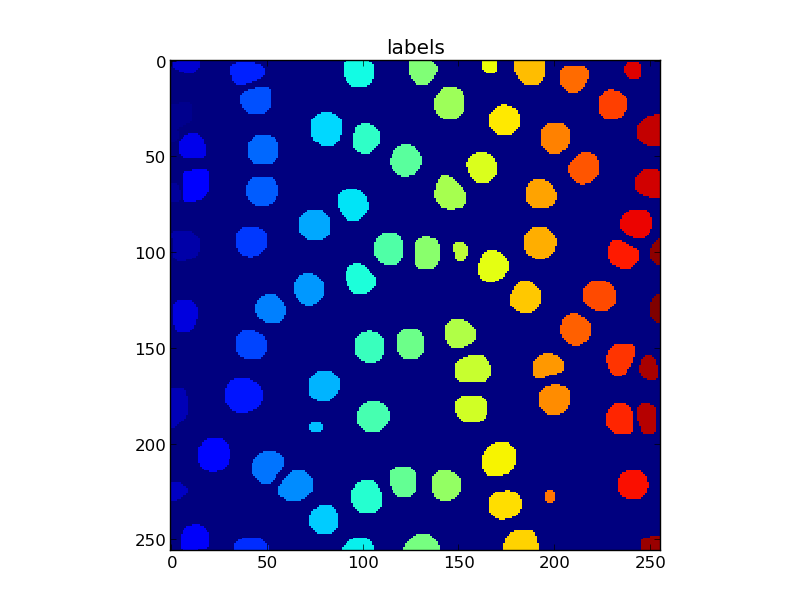}
  }
  \subfigure[Boxes]{
    \label{preproc_1d}
    \includegraphics[width=0.30\textwidth]{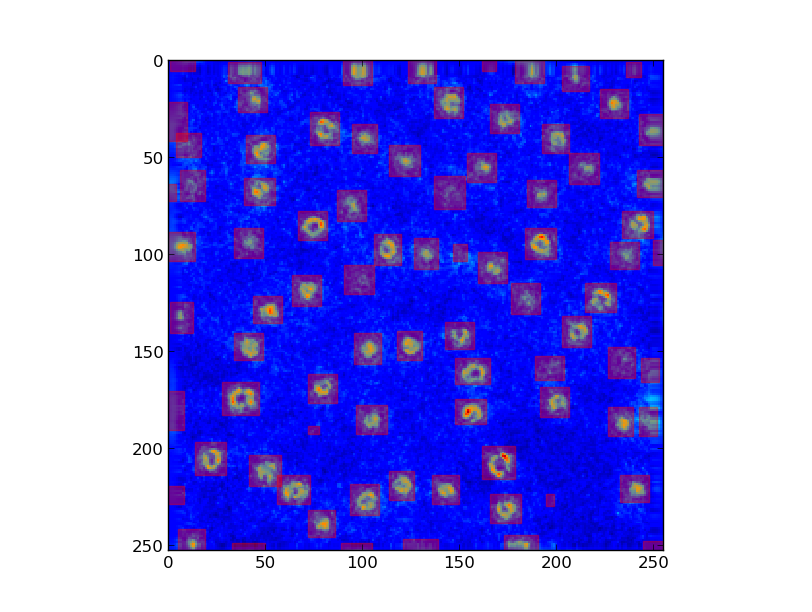}
  }
  \caption{Visualization of the preprocessing procedure, with individual processing steps as described in the main text.}
  \label{preproc}
\end{figure}
Therefore, simple thresholding of the eigenvalues can extract the foreground (cell). For a visualization
of the individual steps of the method, see Figure~\ref{preproc}.
We repeat this procedure at different scales because a large Gaussian kernel strongly suppresses
noise but yields merge errors (i.e., under-segmentation, because mass is aggregated within a larger
neighborhood), while a small Gaussian kernel is sensitive to noise but better preserves the
boundary. Results at different scales have characteristics that are complementary to each other and
combing them produces less false positives and merge errors.

\subsection{Parametric fits}

Our method is optimized for the detection of cell nuclei, 
which are membrane enclosed organelles in eukaryotic cells that contain
most of the cells genetic material. The shape of these objects resembles a deformed ellipsoid.
We argue that we can incorporate this prior knowledge about the
shape by fitting parametric geometric objects such as an $3D$ 
ellipsoid or stack of $2D$ ellipses.

\begin{figure}[h!]
\centering

\subfigure[Parametric fit]{
\includegraphics[width=0.25\textwidth]{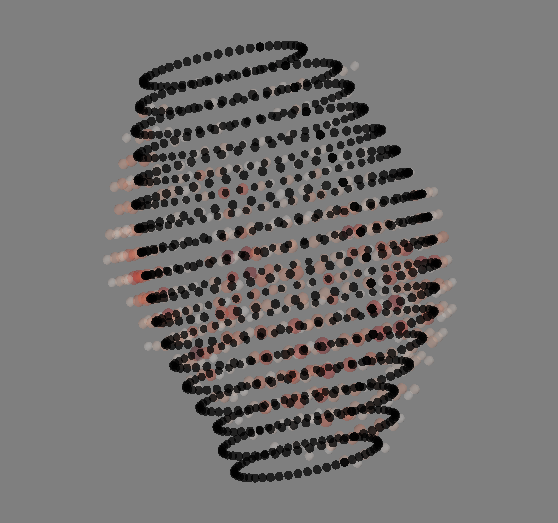}
\label{completeness_1}
}
\subfigure[Layer 15 fit]{
\includegraphics[width=0.25\textwidth]{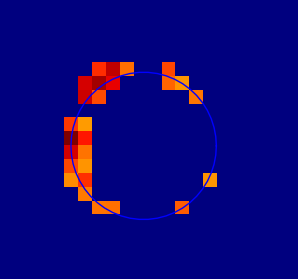}
\label{completeness_2}
}
\subfigure[Layer 20 fit]{
\includegraphics[width=0.25\textwidth]{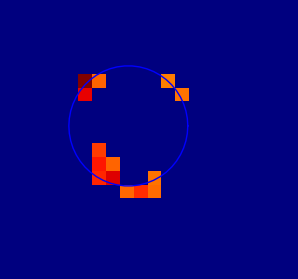}
\label{completeness_3}
}
%\subfigure[MNIST-MTL]{
%\includegraphics[width=0.30\textwidth]{img/supp.pdf}
\label{fig:completeness}
%}
\caption{Subfigure~(a) gives an example of a parametric fit for a volume.
Subfigure~(b) and (c) show two slices through the volume in (a) for
which a good parametric fit is obtained in the face of missing data points.}
\end{figure}
An example how parametric fitting is beneficial in providing robust fits
in the face of missing data points is shown in Figure~\ref{fig:completeness}.
While we limit ourselves to discussing ellipsoid structures for the rest of this paper,
our framework generalizes to other geometric objects such
as splines. % \cite{spline_paper}.

\subsection{Fitting circles}

We start with the simplest parametric object we could use for this task: a circle (one for each layer).
This has obvious limitations, as many nuclei are not perfect circles, but rather correspond to
ellipses in each layer. However, it may be easily derived
and therefore constitutes a good starting point for the description of our method.

The distance of a point $\x \in \mathbb{R}^2,i\in\{1,\dots,n\}$ to a circle with center $\c \in \mathbb{R}^2$ and radius $r \in \mathbb{R}$ is easily computed as
$$d(\c,r,\x) = |\,\|\c - \x\| - r|$$
Finding a circle parametrized by $\c$ and $r$ that minimizes the sum of distances to points $\x_i$
corresponds to solving the following optimization problem:
$$\min_{\c,r} \sum_{i=1}^n L(d(\c,r,\x_i)),$$
where $L$ is a loss function, such as the squared loss or the hinge loss. The choice of loss function
$L$ has important implications on the properties of the fit (e.g., robustness).

\subsection{Fitting ellipses}\label{fit_ellipse}

A class of shapes that allows more flexibility for fitting nuclei in $2D$ are ellipses.
An ellipse in $2D$ can be parametrized by a center point $\c = [c_x, c_y]\transp $ and two radii $\r = [r_x, r_y]\transp $ \cite{Weisstein}.
The points $[x, y]\transp $ on the ellipse centered at $\c$ are then given by the equation
\begin{align}
\frac{(x - c_x)^2}{r_x^2} + \frac{(y - c_y)^2}{r_y^2} = 1.
\end{align}

An alternate parametrization (general conic) is given by
\begin{align}
ax^2 + bxy + cy^2 + dx + ey + f = 0,
\end{align}
%
%TODO: explain how to recover original parameters from this
%
describing an ellipse if $b^2 - 4ac < 0$ \cite{rosin1996assessing}.
Let
\begin{align}
\x &= [x^2, xy, y^2, x, y, 1],\transp \\ \nonumber
\Theta &= [a,b,c,d,e,f]\transp,
\end{align}
then points on the ellipse satisfy $\x\transp  \Theta = 0$. The algebraic distance $f$ of a point $\x$ to the ellipse parametrized by $\Theta$ is defined as:
\begin{align}
f(\x, \Theta) = \x\transp  \Theta .
\end{align}
The algebraic distance  is an approximation of the
Euclidean distance that has the advantage that it is much easier to compute. 
%TODO argue why this is necessary

\paragraph{Avoiding degenerate solutions}

Additional constraints are necessary to avoid degenerate solutions.
In order to avoid the trivial solution $\Theta=0$ and recognizing that any multiple of a solution $\Theta$ represents the same conic, the parameter vector $\Theta$ is constrained in one way or the other \cite{fitzgibbon1999direct}.
Different algorithms for fitting ellipses often only differ in the way they constrain parameters. Many authors suggest $\|a\|^2 = 1$, others $a + c = 1$ or $f = 1$ \cite{fitzgibbon1999direct}.

%TODO make list of different constraints to keep conic an ellipse
%\begin{center}
%\begin{tabular}{lll}
%× & constraint & property\\
%name & × & ×\\
%× & × & ×
%\end{tabular}
%\end{center}
%

\paragraph{Minimizing algebraic distance}

For a general loss-function, we arrive at the following formulation:
\begin{align}
\min_{\Theta} & \sum_{i=1}^N L(\Theta\transp \x_i) \label{general} \\
\textnormal{s.t. }
&\textnormal{solution non-degenerate}\nonumber
\end{align}
Depending on which combination of non-degeneracy constraints and loss function are used, different solvers are needed.

%
%For specific choices of constraints, eigensystems can be solved to obtain a solution.
%
%For example, Bookstein et. al showed that using the quadratic constraint $\|a\|^2 = 1$ to avoid a trivial solution,
%the least-squared solution (i.e. $L_{ls}(r) = r^2$) is given by the rank-deficient generalized eigenvalue system.
%

%
%\subsection{Fitting ellipsoids}
%
%Another geometrical object of interest are ellipsoid,
%which correspond to the generalization of an ellipse
%into the third dimension.
%
%An ellipsiod is paramerized by a center point $c = [c_x, c_y, c_z]\transp $ and three radii $r = [r_x, r_y, r_z]\transp $. The points on a non-rotated ellipse centered at $c$ are described by the equation
%%
%$$\frac{(x - c_x)^2}{r_x^2} + \frac{(x - c_y)^2}{r_y^2} + \frac{(x - c_z)^2}{r_z^2} = 1.$$
%
%TODO: add represtentation for rotated ellipse
%
%The conic parametrization is given by
%%
%$$ax^2 + bxy + cy^2 + dx + ey + f = 0,$$
%
%TODO: how to recover original parameters from this?
%
%describing an ellipse if $b^2 - 4ac < 0$ \cite{rosin1996assessing}.
%%
%Let $\mathbf x = [x^2, xy, y^2, x, y, 1]\transp $ and $\Theta = [a,b,c,d,e,f]\transp $, then $\mathbf x\transp  \Theta = 0$. Let
%%
%$$f(\mathbf x, \mathbf a) = \mathbf \mathbf x\transp  \mathbf a .$$
%
%The function $f$ is called the algebraic distance.
%

\subsection{Robust Loss Function}

\begin{figure}[!ht]
  \centering
  \subfigure[Squared loss]{
    \label{tax_0a}
    \includegraphics[width=0.44\textwidth]{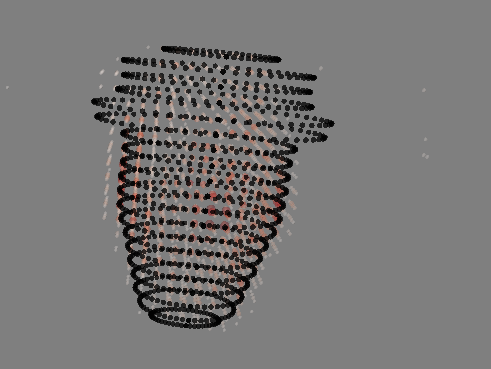}
  }
  \subfigure[$\eps$-insensitive loss ]{
    \label{tax_1a}
    \includegraphics[width=0.44\textwidth]{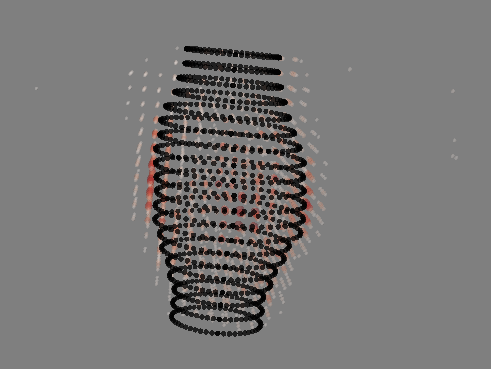}
  }
  \caption{A severe effect of a few outliers for the squared loss is shown in (a). For the same data set, the $\eps$-insensitive loss
       achieves a more robust solution as shown in (b).}
  \label{fig_loss}
\end{figure}

It is well established that the squared loss is particularly prone to outliers, as distance is
penalized quadratically. An example of this sensitivity to outliers is shown in
Figure~\ref{fig_loss}, where a few outliers are sufficient to considerably distort the fit.
We therefore propose to use the $\eps$-insensitive loss function for the problem at hand.
The $\eps$-insensitive loss has its background in the context of Support Vector Regression \cite{Vapnik1995,Scholkopf2002,Smola2004}.
It is also known as dead-zone penalty in other contexts \cite{Boyd} and is often used when a more robust error function is needed.
\begin{align}
L_{\eps}(r) = \bigg\{\begin{array}{cl} |r| - \eps, & \mbox{if } |r| > \eps\\ 0 & \mbox{else.} \end{array}
\end{align}
It has two important properties that make it appealing for the problem at hand. The first is that it does not penalize points that are within
a rim of the stacked ellipsoid. This captures the intuition that the nuclear membrane has a certain thickness and 
we therefore do not want to penalize points that are within the membrane.
Second, the loss is affine (linear minus some offset
that depends on $\eps$). This means that outliers are not penalized as severely
as with the squared loss, yielding a more robust error function.
In the following, we show that although non-differentiable, the
$\eps$-insensitive loss may be expressed in the form of a constrained
optimization problem.
As a first step, we note that $L_{\eps}$ may be written as
\begin{align}
L_{\eps}(r) = \max(|r| - \eps, 0).
\end{align}
Plugging the above loss into Equation (\ref{general}) and neglecting
the non-degeneracy constraints for now, yields the optimization problem:
\begin{align}
\min_{\Theta} & \sum_{i=1}^N \max(|\x\transp  \Theta| - \eps, 0).
\end{align}
We now make use of the fact that $\max(a,b)$ can be expressed
to the smallest upper bound of $a$ and $b$ \cite{Boyd}, i.e.,
\begin{align}
\max(a,b) = \min_c &  \nonumber \\ \label{eq:max_trick}
\textnormal{s.t. }
& a \leq c, \\  \nonumber
& b \leq c.     \nonumber
\end{align}
Furthermore, we exploit that the absolute value may be expressed as the maximum of two linear functions $|r| = \max(r, -r)$.
We use the latter to move $|\x\transp  \Theta|$ from the objective to the
constraints using newly introduced slack variables $s_i$. This gives rise to:
\begin{align}
\min_{\Theta,s_i} & \sum_{i=1}^N \max(s_i - \eps, 0)  \nonumber \\ %\label{eq:reg_dual}\\
\textnormal{s.t. }
& \x_i\transp  \Theta \leq s_i, \\  \nonumber
-& \x_i\transp  \Theta \leq s_i   \nonumber
\end{align}
Using the same scheme, the other $\max$ is moved to the constraints using Equation (\ref{eq:max_trick}),
introducing variables $t_i$. We arrive at:
\begin{align}
\min_{\Theta,s_i,t_i} \sum_{i=1}^N & t_i \nonumber \\ \label{eq:robust}
\textnormal{s.t. }
  \x_i\transp  \Theta &\leq s_i, \\ \nonumber
 -\x_i\transp  \Theta &\leq s_i, \\ \nonumber
         s_i - \eps &\leq t_i, \\ \nonumber
                 0 &\leq t_i.     \nonumber
\end{align}

%TODO: say some final words

\subsection{Graph-regularization}

\begin{figure}[H]
  \centering
  \subfigure[No regularization]{
    \label{tax_0b}
    \includegraphics[width=0.44\textwidth]{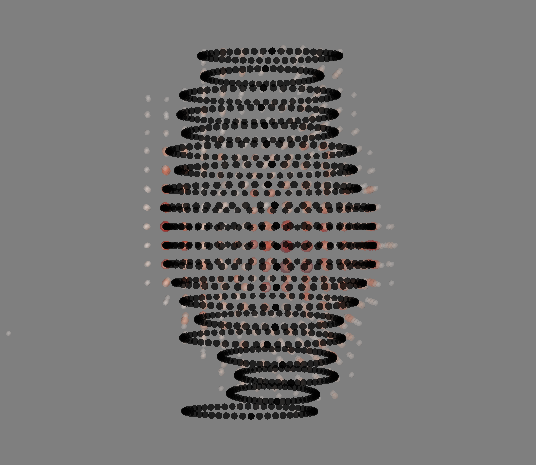}
  }
  \subfigure[Graph regularization]{
    \label{tax_1b}
    \includegraphics[width=0.44\textwidth]{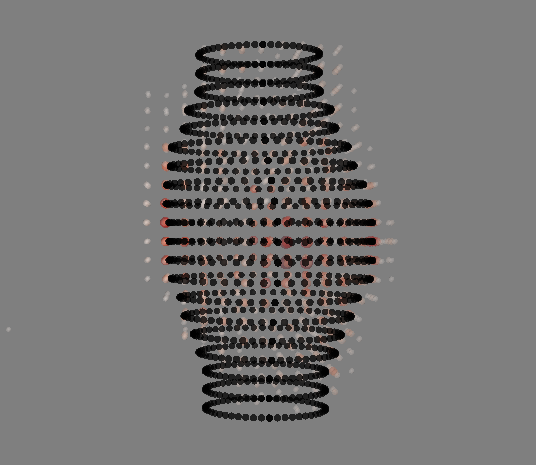}
  }
  \caption{Example of cell fitting in each z-layer independently (a) and with graph regularization (b). 
       In the top and bottom layers of the nucleus, the non-regularized $2D$ fits differ greatly between
       layers as only a few data points are available for each layer. By coupling layers via Multitask Regularization,
       we achieve a smooth fit.}
  \label{fig_reg}
\end{figure}

To share information of the ellipse fitting across the z-layers , we propose to
jointly fit ellipses in all layers and penalize differences between parameter vectors
of neighboring layers by means of regularization term $R$:.
\begin{align}
R(\Theta_1,\dots,\Theta_N) = \sum_{i=1}^{N-1} \|\Theta_i - \Theta_{i+1}\|_p, \label{reg_neighbor}
\end{align}
where $\|a\|_p$ is the $p$-norm. 
The effect of this smoothing is shown in Figure~\ref{fig_reg}.
This smoothness regularizer is a special case of a general graph-regularizer, which is often used in
the context of Multitask Learning \cite{evgeniou2006learning,Widmer2012}, as edges only exist between
neighboring layers.
Note that in the above formulation, we have not settled on a particular norm, however in the
following we will instantiate to the $L_1$-norm.

\subsection{Linear Program (LP) Formulation}

We now start putting all pieces together to obtain the final optimization problem.  Starting from
Equation (\ref{eq:robust}), we add the graph-regularizer from Equation (\ref{reg_neighbor}) to the
mix.

Note that to avoid the trivial solution, we add additional constraints as discussed in
Section~\ref{fit_ellipse}.  Here, we use $\Theta_{i,1} + \Theta_{i,3} = 1$ (i.e., $a + c = 1$).
\begin{align}
\min_{\Theta_i,s_i,t_i} \sum_{t=1}^M \sum_{i=1}^{N_t} & t_{t,i} + \sum_{t=1}^{M-1} |\Theta_i - \Theta_{i+1}|_1  \nonumber \\ \nonumber %\label{eq:reg_dual}\\
\textnormal{s.t. }
  \x_i\transp  \Theta &\leq s_i, \\ \nonumber
 -\x_i\transp  \Theta &\leq s_i, \\ \nonumber
      s_i - \eps &\leq t_i, \\ \nonumber
              0 &\leq t_i, \\ \nonumber
\Theta_{i,1} + \Theta_{i,3} &= 1.     \nonumber
\end{align}

Again, using the fact that $|a| = max(a, -a)$, we push the graph-regularizer to the constraints:
\begin{align}
\min_{\Theta_i,s_i,t_i} \sum_{i=1}^M \sum_{j=1}^{N_i} & t_{i,j} + \sum_{i=1}^{M-1} \sum_{j=1}^{D} u_{ij} \nonumber \\ \nonumber %\label{eq:reg_dual}\\
\textnormal{s.t. }
  \Theta_{i,j} - \Theta_{i,j} &\leq u_{i,j} \forall i \in [1,M-1], j \in [1,D], \\ \nonumber
  -\Theta_{i,j} + \Theta_{i,j} &\leq u_{i,j} \forall i \in [1,M-1], j \in [1,D], \\ \nonumber
  \x_i\transp  \Theta &\leq s_i, \\ \nonumber
 -\x_i\transp  \Theta &\leq s_i, \\ \nonumber
         s_i - \eps &\leq t_i, \\ \nonumber
              0 &\leq t_i, \\ \nonumber
\Theta_{i,1} + \Theta_{i,3} &= 1.     \nonumber
\end{align}
The above problem consists of a linear objective and linear constraints and can therefore be solved
with a linear program solver. We used the freely available GNU Linear Programing Toolkit to solve
the above optimization problem (\url{http://www.gnu.org/software/glpk/glpk.html}).

\begin{figure}[!h]
  \centering
  \includegraphics[width=0.98\textwidth]{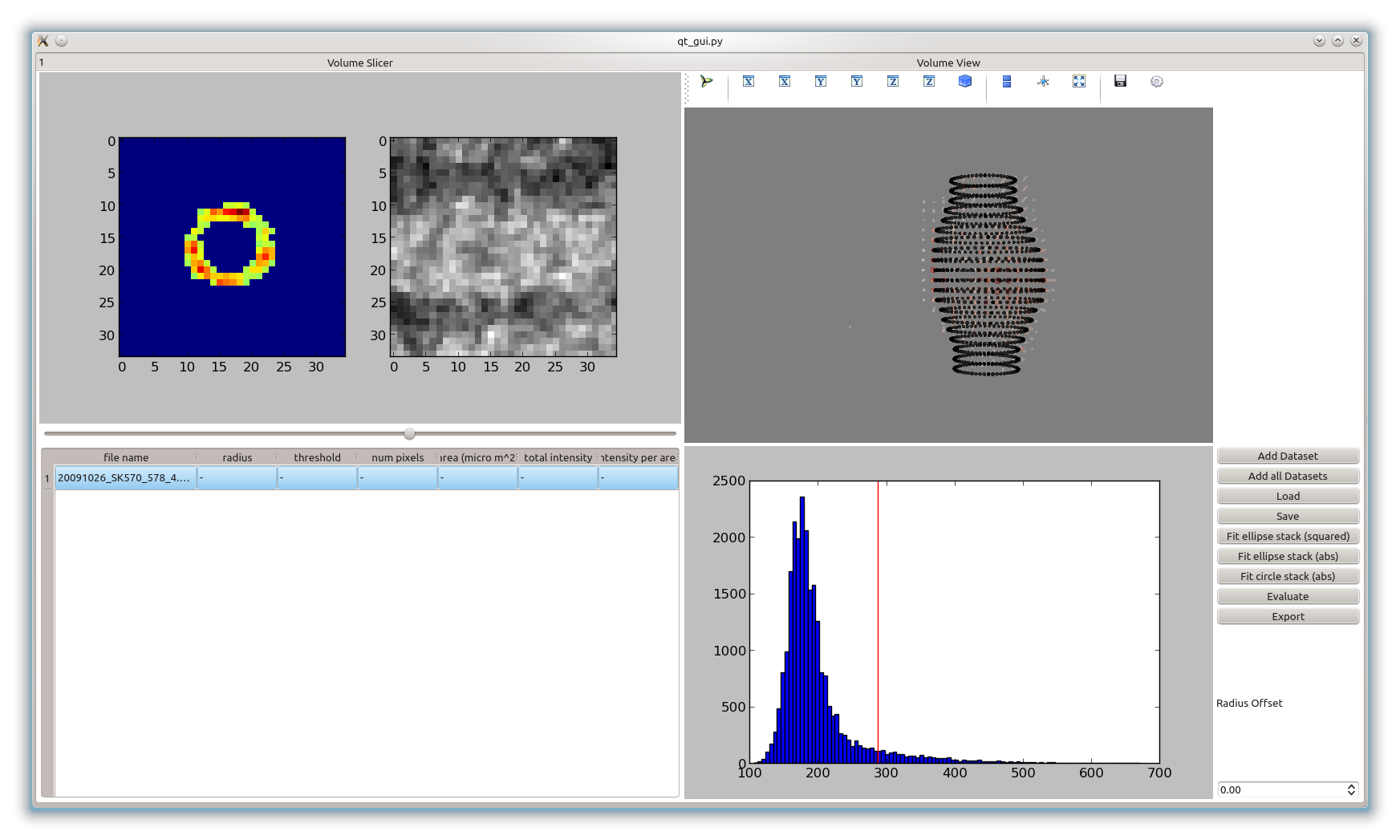}
  \caption{Graphical User Interface for rapidly performing experiments.}
  \label{fig_gui}
\end{figure}

For ease of use, we provide a graphical user interface (GUI) as shown in Figure~\ref{fig_gui}.

\section{Experiments}

\begin{figure}[h!]
\centering

\subfigure[low noise]{
\includegraphics[width=0.31\textwidth]{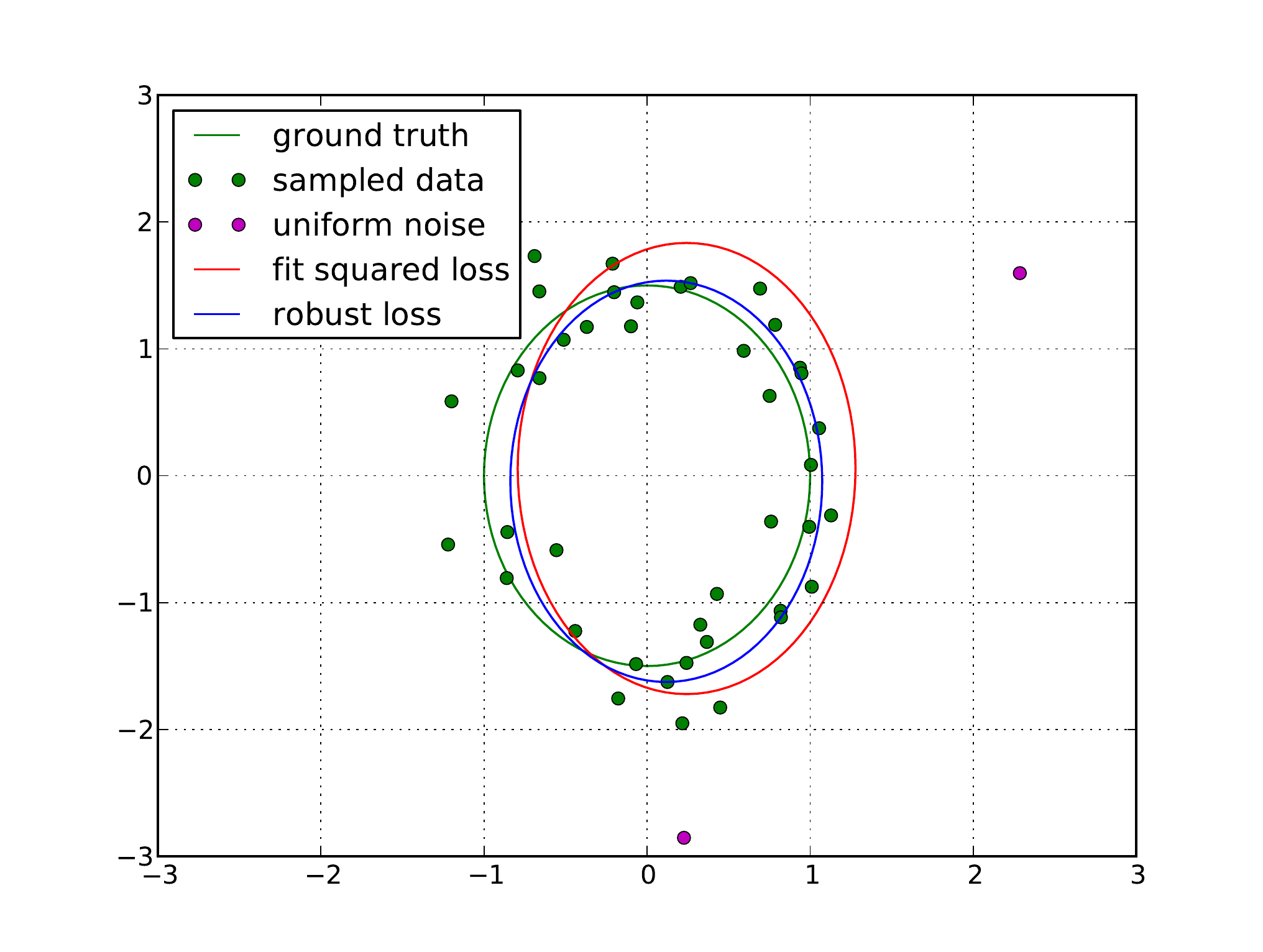}
\label{robust_stage1}
}
\subfigure[medium noise]{
\includegraphics[width=0.31\textwidth]{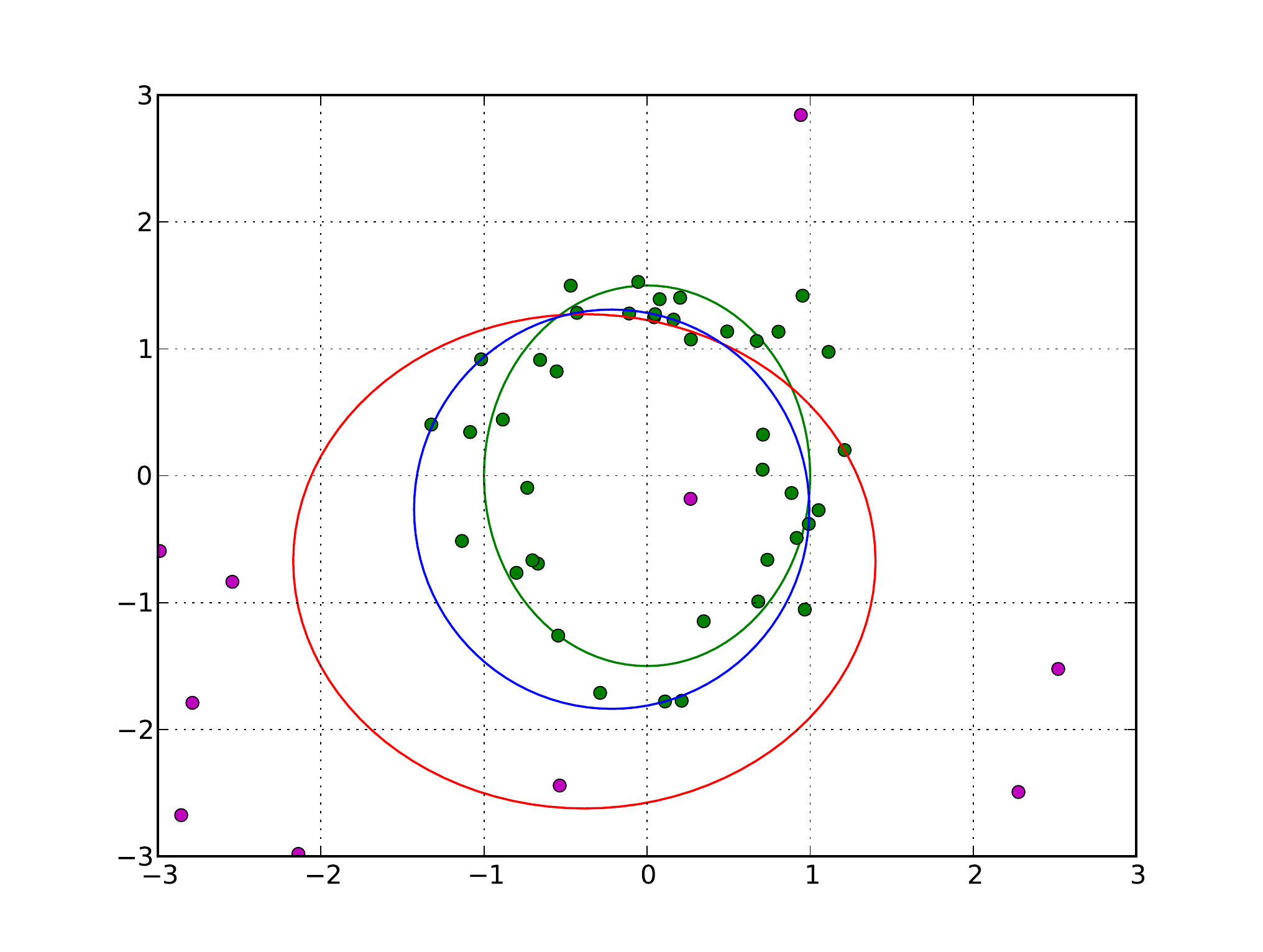}
\label{robust_stage2}
}
\subfigure[high noise]{
\includegraphics[width=0.31\textwidth]{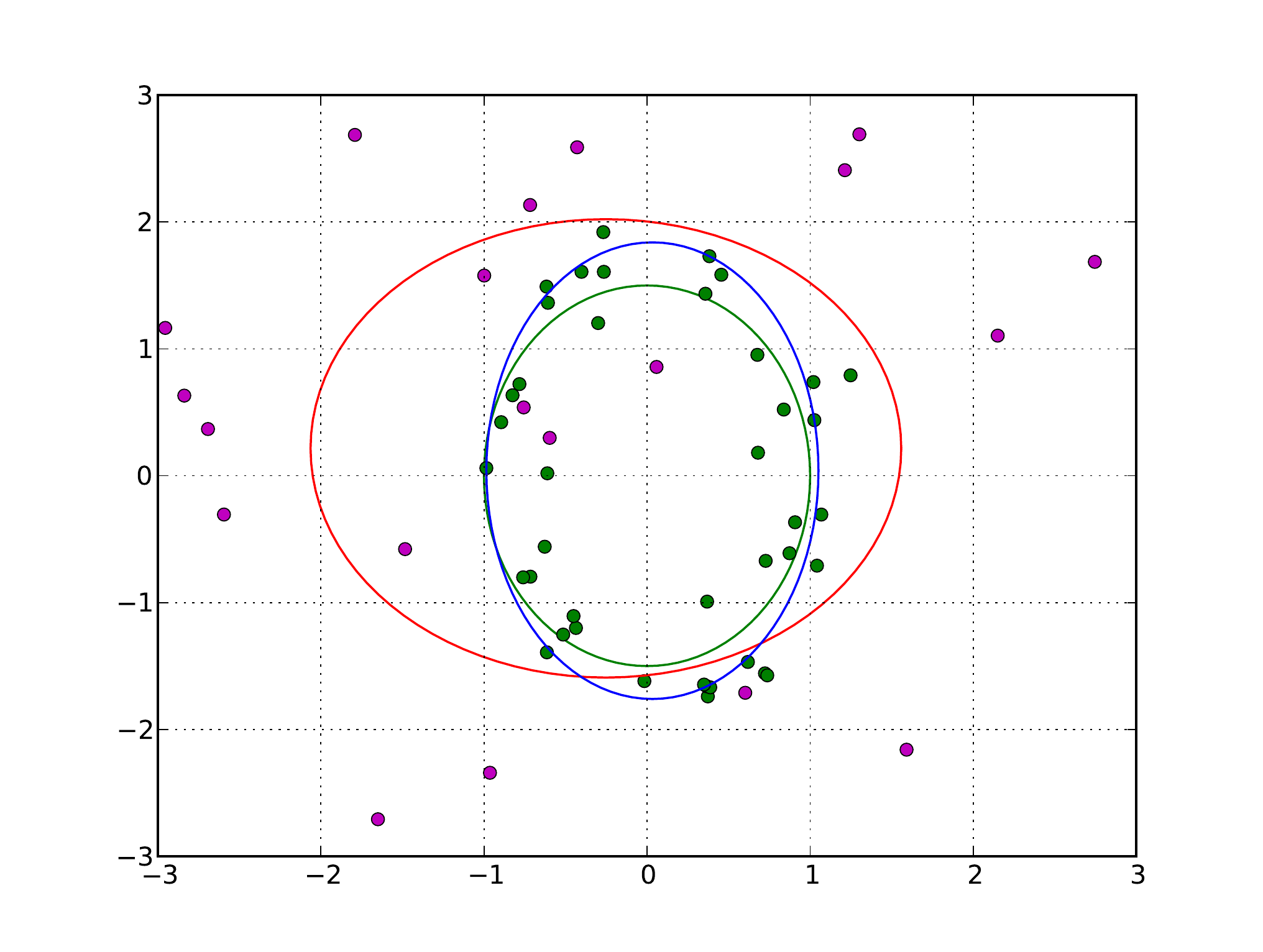}
\label{robust_stage3}
}
\label{fig:robust}
%}0
\caption{Fits for different noise regimes. The dots that are sampled from the noise distribution are shown in (purple) and the ones sampled from the underlying ellipse (green) are shown in (black). The true squared loss fit is shown in (red) and the one using the robust loss is shown in (blue).}
\end{figure}

\begin{figure}[ht]
\begin{center}
  \includegraphics[width=0.9\textwidth]{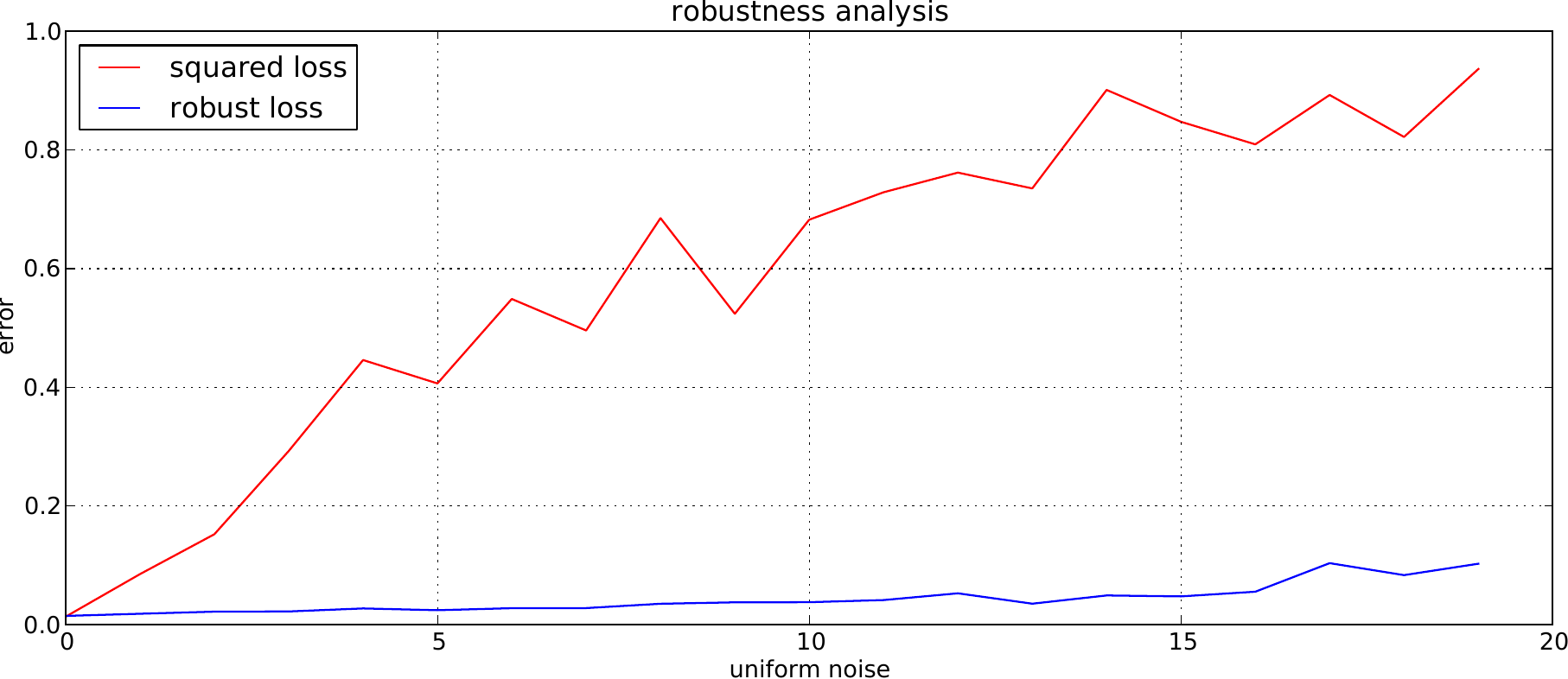}
\caption{Comparison of the fitting error of the squared and robust loss for different mixtures of the distributions of the ellipsoid and the noise.}
\label{fig:robust_eval}
\end{center}
\end{figure}

\paragraph{Robustness Analysis}
To compare robust loss and squared loss in the context of parametric fits, we set up an experiment
using synthetic data. For this, we first sampled $n$ points $\x_1, ..., \x_n$ from an ellipse
parametrized by $\Theta$.  Next, uniformly distributed points were sampled in the interval $[-3, 3]$
to simulate random noise and contaminations.  Based on all sampled points, two fits were obtained,
one using the squared loss and one using the robust loss.  Examples of these fits are shown in
\ref{robust_stage1}, \ref{robust_stage2} and \ref{robust_stage3}.  A systematic comparison of the
two losses is shown in Figure~\ref{fig:robust_eval}, where the error with respect to the ground
truth (i.e., $\|\Theta_{fit} - \Theta\|$) is shown as a function of the number of uniformly sampled
points. We observe that the error increases much later when using the robust loss function as
opposed to using the squared loss.

\begin{figure}[h!]
\centering

\subfigure[Microscope software]{
\includegraphics[width=0.30\textwidth]{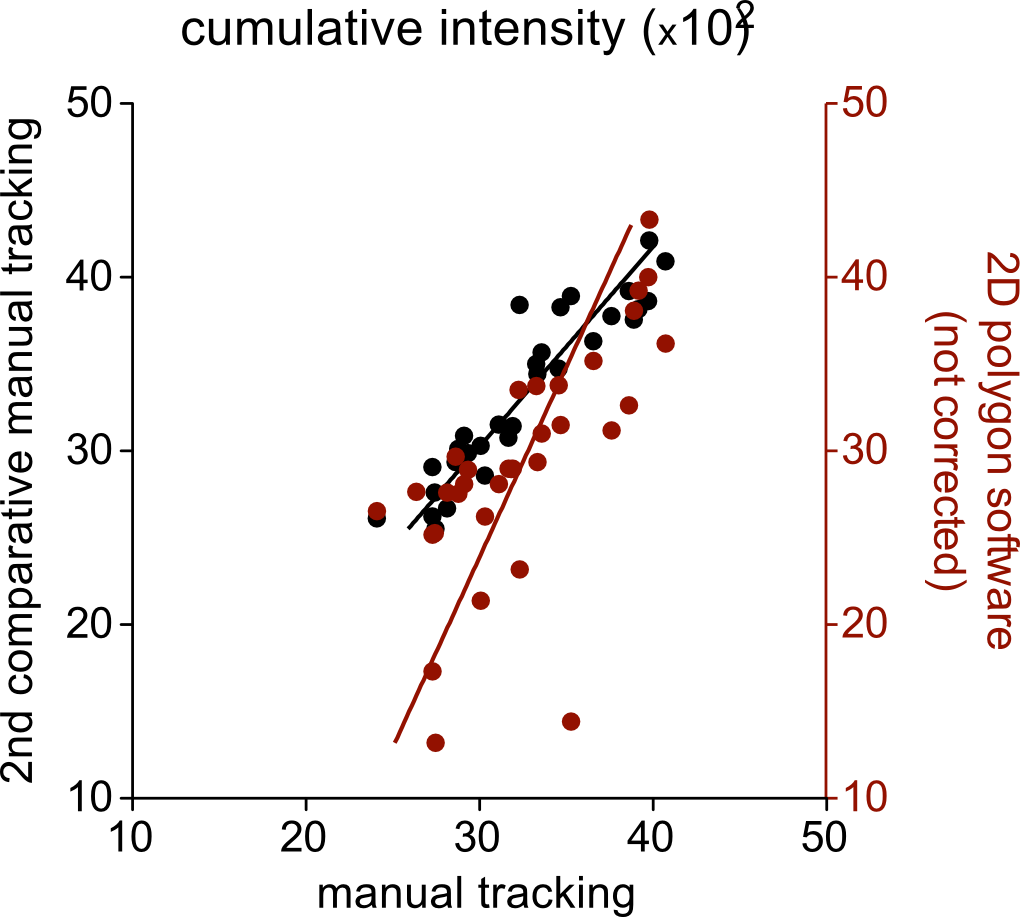}
\label{result_1}
}
\subfigure[Our approach]{
\includegraphics[width=0.30\textwidth]{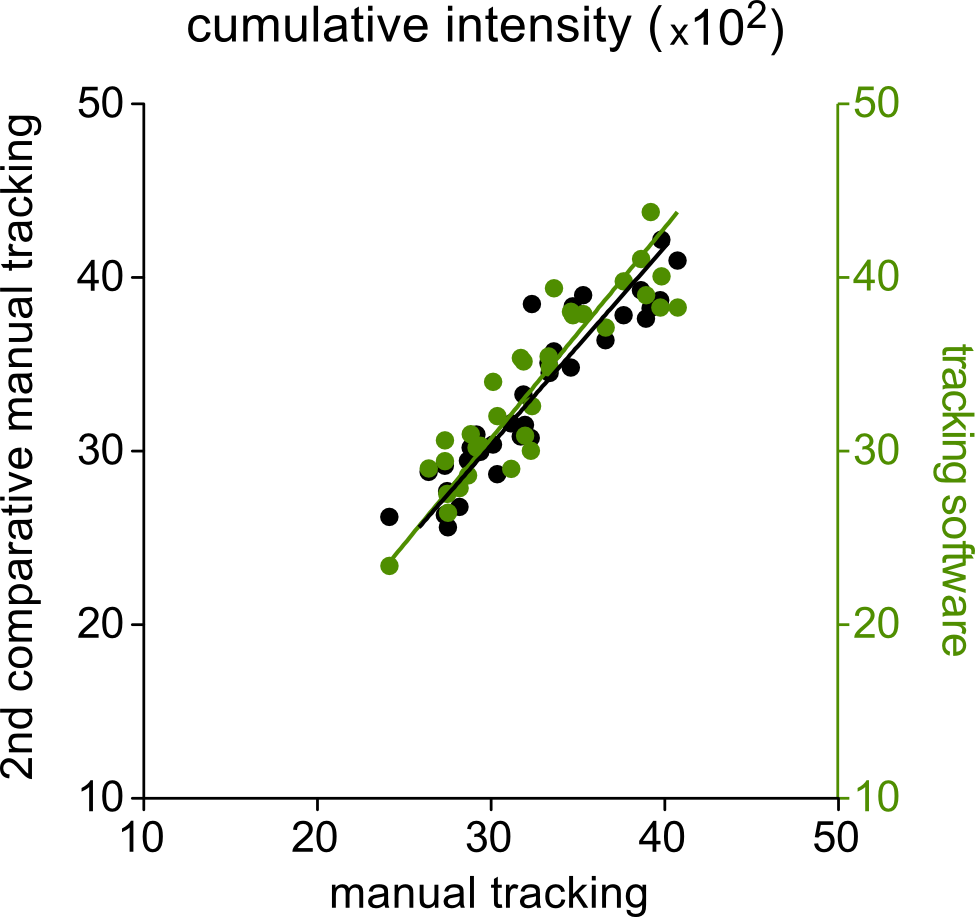}
\label{result_2}
}
\subfigure[Time gain]{
\includegraphics[width=0.30\textwidth]{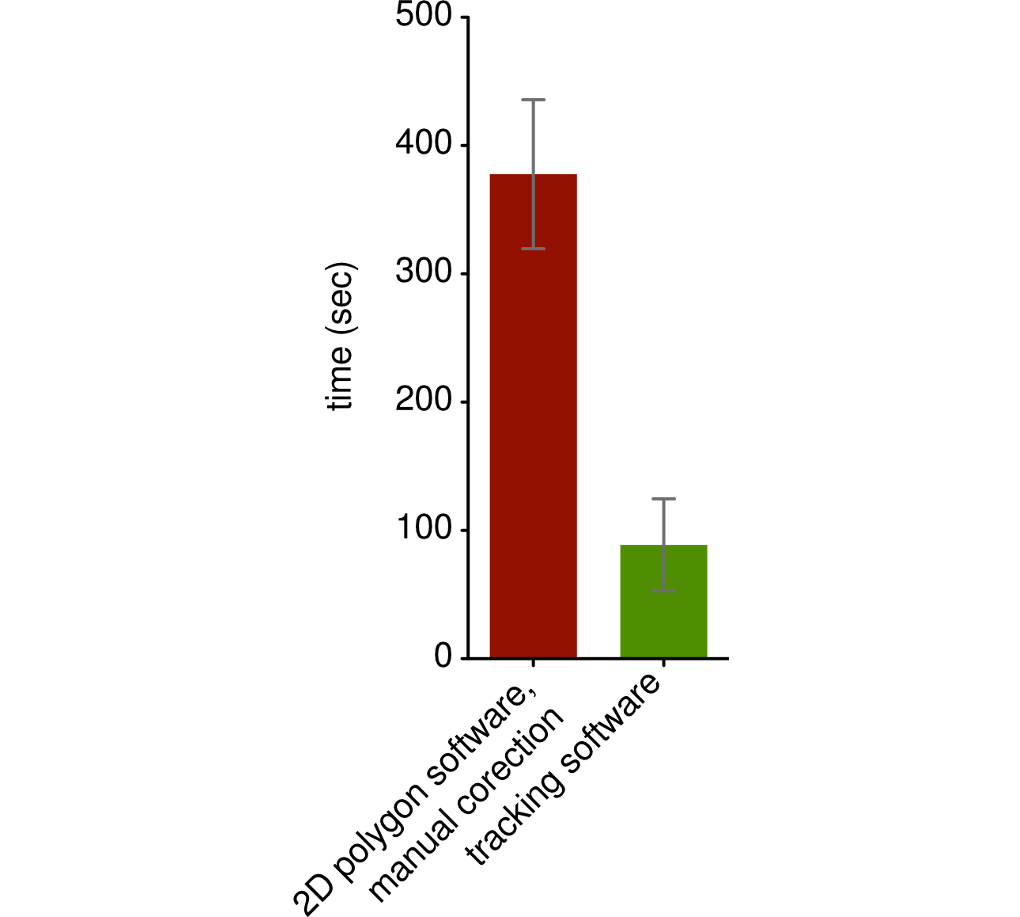}
\label{result_3}
}
%\subfigure[MNIST-MTL]{
%\includegraphics[width=0.30\textwidth]{img/supp.pdf}
%\label{fig:subfigA3}
%}
\caption{\label{fig:results}Results on real data comparing the $2D$ polygon fit 
by the microscope manufacturer (a) to our approach (b). Taking into account manual post processing,
the total time taken for an experiment is shown in (c).}
\end{figure}

\paragraph{Evaluation in Practice}
In order to evaluate the quality of our fits on real data, we compared them to manually curated
segmentations obtained using the software of the microscope manufacturer.  The results are shown in
Figure \ref{result_2}. We observe that our approach has an almost perfect correlation to the
manually curated ground truth, while the existing microscope software shows a considerable deviation
(see Figure \ref{result_1}).
As we are ultimately interested in allowing for high-throughput experiments, we quantified the time
taken to perform an experiment using our approach and the microscope software and report a large
speed-up.  Note that the experiment was conducted by fitting each cell individually.  Taking into
account the proposed preprocessing pipeline and setting up batch processing will almost entirely automate the
procedure.

\section{Conclusion}
We have presented a tool for parametric fitting of cell-like objects in fluorescence microscopy
images. We have shown that using modern machine learning and computer-vision techniques, we can
greatly speed-up the experimental process and outperform existing software.
Combined with an easy-to-use user interface, our tool enables biologists to perform truly
high-throughput quantitative experiments in fluorescence microscopy~\cite{heinrich}.

\paragraph{Acknowledgements} 
We gratefully acknowledge core funding from the Sloan-Kettering Institute (to G.R.), 
from the Ernst Schering foundation (to S.H.) and from the Max Planck Society (to G.R. and S.H.).
Part of this work was done while C.W. P.D. and G.R. were at the Friedrich Miescher Laboratory of the Max Planck Society 
and while C.W. was at the Machine Learning Group at TU-Berlin.

\bibliography{cell_fitting_paper}

\newcommand{\etalchar}[1]{$^{#1}$}
\begin{thebibliography}{WKGR12}

\bibitem[{Boy}04]{Boyd}
L.~{Boyd, S.P. and Vandenberghe}.
\newblock {\em {Convex optimization}}.
\newblock Cambridge University Press, 2004.

\bibitem[EMP05]{evgeniou2006learning}
T.~Evgeniou, C.A. Micchelli, and M.~Pontil.
\newblock {Learning multiple tasks with kernel methods}.
\newblock {\em Journal of Machine Learning Research}, 6(1):615--637, 2005.

\bibitem[FPF99]{fitzgibbon1999direct}
A.~Fitzgibbon, M.~Pilu, and R.B. Fisher.
\newblock {Direct least square fitting of ellipses}.
\newblock {\em Pattern Analysis and Machine Intelligence, IEEE Transactions
  on}, 21(5):476--480, 1999.

\bibitem[GW08]{Gonzalez2008Digital}
R.~C. Gonzalez and R.~E. Woods.
\newblock {\em {Digital Image Processing}}.
\newblock Prentice Hall, Upper Saddle River, N.J., 2008.

\bibitem[HGK{\etalchar{+}}13]{heinrich}
S.~Heinrich, E.M. Geissen, J.~Kamenz, S.~Trautmann, C.~Widmer, P.~Drewe,
  M.~Knop, N.~Radde, J.~Hasenauer, and S.~Hauf.
\newblock {Determinants of robustness in spindle assembly checkpoint
  signalling}.
\newblock {\em Nature Cell Biology (in press)}, 2013.

\bibitem[LKWH12]{Lou2012Learning}
X.~Lou, U.~Koethe, J.~Wittbrodt, and F.~A. Hamprecht.
\newblock {Learning to Segment Dense Cell Nuclei with Shape Prior}.
\newblock In {\em CVPR}, 2012.

\bibitem[Ros96]{rosin1996assessing}
P.L. Rosin.
\newblock {Assessing error of fit functions for ellipses}.
\newblock {\em Graphical models and image processing}, 58(5):494--502, 1996.

\bibitem[SS02]{Scholkopf2002}
B~Sch\"{o}lkopf and A~J Smola.
\newblock {\em {Learning with Kernels}}, volume~64 of {\em Adaptive Computation
  and Machine Learning}.
\newblock MIT Press, 2002.

\bibitem[SS04]{Smola2004}
A.J. Smola and B.~Sch\"{o}lkopf.
\newblock {A tutorial on support vector regression}.
\newblock {\em Statistics and Computing}, 14(3):199--222, 2004.

\bibitem[Vap95]{Vapnik1995}
V~N Vapnik.
\newblock {\em {The Nature of Statistical Learning Theory}}, volume~8 of {\em
  Statistics for Engineering and Information Science}.
\newblock Springer, 1995.

\bibitem[Wei]{Weisstein}
Eric~W. Weisstein.
\newblock {Ellipse -- from Wolfram MathWorld}.

\bibitem[WKGR12]{Widmer2012}
C.~Widmer, M.~Kloft, N.~G\"{o}rnitz, and G.~R\"{a}tsch.
\newblock {Efficient Training of Graph-Regularized Multitask SVMs}.
\newblock In {\em {ECML2012}}, 2012.

\end{thebibliography}

\end{document}